\documentclass[letterpaper, 10 pt, conference]{ieeeconf}  

\IEEEoverridecommandlockouts                              

\usepackage{graphics} 
\usepackage{mathrsfs}
\usepackage{times}
\usepackage{epsfig}
\usepackage{amsmath}
\usepackage{amssymb}
\usepackage{threeparttable}
\usepackage{booktabs}
\usepackage{amsfonts, amssymb}
\usepackage{multirow,bigdelim}
\usepackage{subfigure}
\usepackage{indentfirst}
\usepackage{caption}
\usepackage{geometry}

\usepackage[pagebackref=true,breaklinks=true,letterpaper=true,colorlinks,bookmarks=false]{hyperref}

\title{\LARGE \bf
FlowMOT: 3D Multi-Object Tracking by Scene Flow Association
}

\author{Guangyao Zhai$^{1}$, Xin Kong$^{1}$, Jinhao Cui$^{1}$, Yong Liu$^{1,2,*}$ and Zhen Yang$^{3}$\\
		\thanks{$^{1}$Guangyao Zhai, Xin Kong, Jinhao Cui and Yong Liu are with the Institute of Cyber-Systems and Control, Zhejiang University, Hangzhou 310027, P. R. China.}%
		\thanks{$^{2}$Yong Liu is with the State Key Laboratory of Industrial Control Technology, Zhejiang University, Hangzhou 310027, P. R. China (*Yong Liu is the corresponding author, Email: \url{yongliu@iipc.zju.edu.cn}).}
		\thanks{$^{3}$Zhen Yang is with the Noah's Ark Laboratory, Huawei Technologies Co. Ltd., Shanghai 200127, P. R. China.}%
}

\begin{document}

\maketitle
\thispagestyle{empty}
\pagestyle{empty}

\begin{abstract}
        Most end-to-end Multi-Object Tracking (MOT) methods face the problems of low accuracy and poor generalization ability. Although traditional filter-based methods can achieve better results, they are difficult to be endowed with optimal hyperparameters and often fail in varying scenarios. To alleviate these drawbacks, we propose a LiDAR-based 3D MOT framework named FlowMOT, which integrates point-wise motion information with the traditional matching algorithm, enhancing the robustness of the motion prediction. We firstly utilize a scene flow estimation network to obtain implicit motion information between two adjacent frames and calculate the predicted detection for each old tracklet in the previous frame. Then we use Hungarian algorithm to generate optimal matching relations with the ID propagation strategy to finish the tracking task. Experiments on KITTI MOT dataset show that our approach outperforms recent end-to-end methods and achieves competitive performance with the state-of-the-art filter-based method. In addition, ours can work steadily in the various-speed scenarios where the filter-based methods may fail.

\end{abstract}

\section{Introduction}

The task of Multi-Object Tracking (MOT) has been a long-standing and challenging problem, which aims to locate objects in the video and assign a consistent ID for the same instance. Many vision applications, such as autonomous driving, robot collision prediction, and video face alignment, require MOT as their crucial component technology. Recently, MOT has been fueled along with the development of research concerned with 3D object detection.
\begin{figure}[t]
	\begin{center}
		\includegraphics[width=1.0\linewidth]{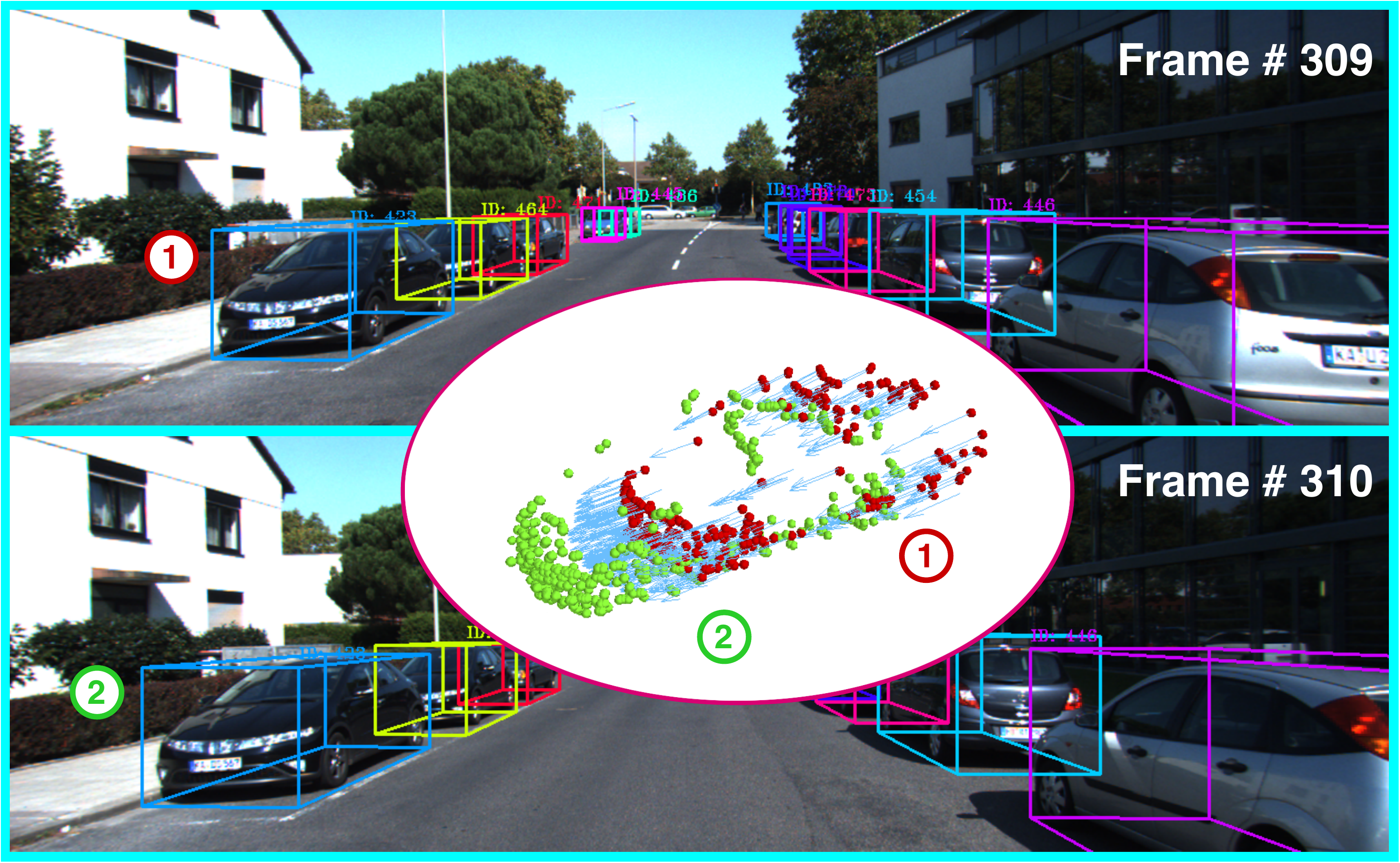}
	\end{center}
	\caption{A glimpse of FlowMOT results in sequence 01. Red 1 and Green 2 stand for the same chosen car in two adjacent frames. Blue arrows represent the point-wise scene flow belonging to the car.}
	\label{fig:method_compare}
	\vspace{-4mm}
\end{figure}

The researches on 3D MOT have attracted widespread attention in the vision community. Weng \textit{et.al.}~\cite{weng20203d} presented a Kalman Filter-based 3D MOT framework based on LiDAR point clouds and proposed a series of evaluation metrics for such task. Although impressive performance both on accuracy and operation speed, a notable drawback is that such a method only focuses on the bounding boxes of the detection results but ignores the internal correlation of point clouds. Besides, Kalman Filter requires frequent hyperparameter tuning due to hand-crafted motion models, which are sensitive to the attributes of tracked objects, like speed and categories (e.g., optimal settings for Car are various from the ones for Pedestrian). In the following work based on Kalman Filter, Chiu \textit{et.al.}~\cite{chiu2020probabilistic} has further improved the performance by implementing a hyperparameter-learning method based on collecting distribution information of the dataset. However, each dataset has its own characteristics (e.g., the condition of sensors and collection frequencies vary in different datasets), theoretically making this method a paucity of generalization across various datasets. As for current end-to-end approaches, they normally obtain a restrained accuracy, and the explainability is not apparent enough.

Inspired by using the optical flow of the 2D object pixel-level segmentation results to generate tracklets in \cite{luiten2020track}, we realize that scene flow reflects the point-wise 3D motion consistency and has the potential ability to tackle the above issues by integrating with traditional matching algorithms. We follow the track-by-detection paradigm and propose a hybrid motion prediction strategy based on the contribution of point-wise consistency. We use the estimated scene flow to calculate the object-level movement to make it have tracklet-prediction ability. Then, through some match algorithms (such as the Hungarian algorithm~\cite{kuhn1955hungarian}), we can get the matching relationship between old tracklets in the previous frames and all new detection results in subsequent ones to update the tracking outcomes. We use the proposed \textit{Motion Estimation} and \textit{Motion Tracking} modules, taking advantage of scene flow in tracklet prediction instead of Kalman Filter, which saves the trouble of adjusting numerous hyperparameters and achieves the intrinsic variable motion model construction through the whole process. Experiments show that our framework can maintain satisfactory performance, especially in challenging various-speed scenarios where traditional methods may fail due to improper tuning. 

We expect our framework can serve as a prototype for drawing researchers' attention to explore related motion-based methods. Our contributions are as follows:

\begin{itemize}
\item We propose a 3D MOT framework with scene flow estimation named FlowMOT, which, to the best of our knowledge, firstly introduces learning-based scene flow estimation into 3D MOT.

\item By integrating the learning-based motion prediction module to update the object-level predictions in 3D space, we avoid the trouble of repeatedly tuning the hyperparameters and constant motion model issues inherent in the traditional filter-based methods.

\item Experiments on the KITTI MOT dataset show that our approach achieves competitive performance against state-of-art methods and can work in challenging scenarios where the traditional methods usually fail.
\end{itemize}

\section{Related Work}

In recent years, various methods have been explored for MOT, but in terms of methodology, so far, most MOT solutions are based on two paradigms, namely track-by-detection and joint detection and tracking.

\smallskip
\noindent
\textbf{Track-by-detection.}
Most tracking methods are developed based on this paradigm, which can be attributed to prediction, data association, and update. As the most critical part, data association has attracted tons of researchers to this field. In the traditional methods, SORT~\cite{bewley2016simple} used Kalman Filter for object attribute prediction and then used Hungarian algorithm~\cite{kuhn1955hungarian} for data association. Weng et al.~\cite{weng20203d} applied this paradigm to the tracking of 3D objects, depending on the Hungarian algorithm as well, and propose a new 3D MOT evaluation standard similar to the 2D MOT but overcome its drawbacks. This filter-based tracking framework can achieve accurate results with real-time performance, but its disadvantage is evident in the meantime. To begin with, it needs to adjust hyperparameters in the covariance matrixes to achieve promising results. Secondly, its motion model is too constant to generalize across different circumstances. However, in practical usage, application scenes are diversified, and it is not easy to unify a standard motion model that can adapt to all conditions. CenterPoint~\cite{yin2020center} proposed a BEV-based detector and used information collected in advance (e.g., speed and orientation) to associate the center of the object by the Hungarian-based closest-point matching algorithm. Recently, due to the robust feature extraction ability of deep learning, more and more learning-based methods for data association has been explored. \cite{frossard2018end} proposed the Deep Structured Model (DSM) method based on hinge loss, which is entirely end-to-end. mmMOT~\cite{zhang2019robust} used off-the-shelf detectors, including images and point clouds for data association through multi-modal feature fusion and learning of the adjacency matrix between objects. The above two methods both use linear programming~\cite{schulter2017deep} and belong to offline methods in tracking research. \cite{weng2020gnn3dmot} proposed to interact with appearance features through a Graph Neural Network, thus helping the association and tracking process. JRMOT~\cite{shenoi2020jrmot} was designed as a hybrid framework combining traditional and deep learning methods. It also used camera and LiDAR sensors, but it is an online method and improves the traditional data association part by fusing the appearance based on the two modalities -- IoU and appearance are selected through training to build a better cost matrix then perform JPDA procedure. However, the 3D appearance used in this method does not have the object's representative features (2D is available because it uses the ReID technique~\cite{zhang2017alignedreid}). A proper exploration of the 3D appearance extraction will undoubtedly help this performance in the future.

Our method follows this paradigm. In addition, we use scene flow estimation to extract 3D motion consistent information for prediction instead of Kalman Filter to avoid hyperparameter adjustments. \cite{lenz2011sparse} had proposed to use sparse scene flow for moving object detection and tracking, but using image information to calculate geometric leads to relatively lower accuracy. In contrast, we benefit from the robustness and higher accuracy of the learning-based 3D scene flow to achieve better results.

\smallskip
\noindent
\textbf{Joint detection and tracking.}
As the detection task and the tracking task are positively correlated, another paradigm different from track-by-detection is created, allowing detection and tracking to refine each other or integrate the two tasks for a multi-task completion. \cite{feichtenhofer2017detect} introduced a siamese network with current and past frame inputs that is used to predict the offsets of the bounding boxes to complete the tracking work, using the Viterbi algorithm to maximize the output of a proposed score function to optimize the detection result. MOTSFusion~\cite{luiten2020track} faced the Multi-Object Tracking and Segmentation~\cite{Voigtlaender2019CVPR} by using multi-modality data to generate tracklets and combines the reconstruction task to optimize and integrate the final track results by detecting the continuity between tracks, thereby recovering missed detection. Tracktor~\cite{bergmann2019tracking} utilized a previous robust detector and adopts the strategy of ID propagation to achieve tracking without complex algorithms. Inspired by this idea, CenterTrack~\cite{zhou2020tracking} uses the two frames of images and the heat map of the previous frame as input to the network to predict the bounding boxes and offsets of the next frame achieving the integration of the two tasks.

\begin{figure*}[t]
	\begin{center}
		\includegraphics[width=1.0\linewidth]{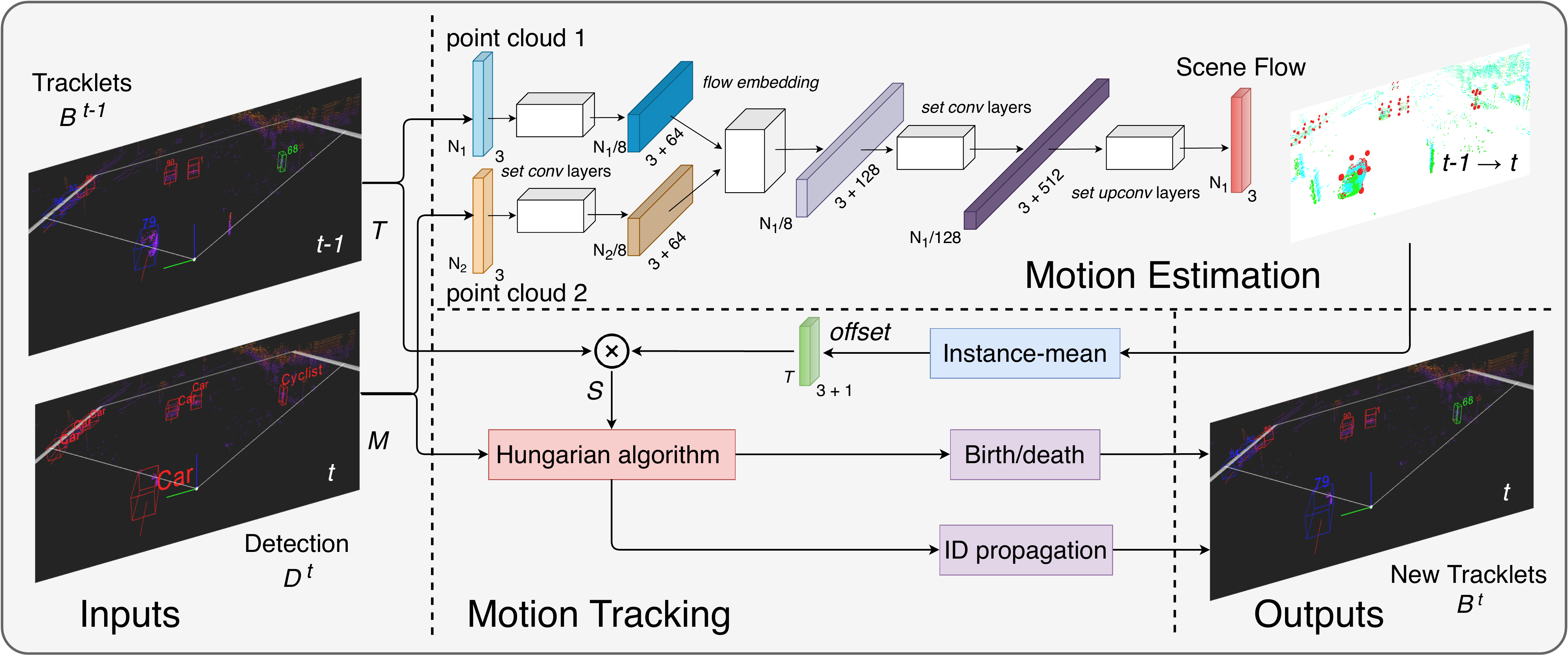}
	\end{center}
	\caption{The FlowMOT architecture for 3D MOT task. The whole framework consists of three components. For the inputs, we use two adjacent ground-regardless point clouds along with detection results of $t$ frame and the tracklets of $t-1$ frame. Motion Estimation module is used to extract 3D motion information from point clouds expressing in the scene flow format. Next step, Motion Tracking module transfers the estimated scene flow into a series of \textit{offset} to help the data association procedure. Finally we use ID propagation plus birth/death check to accomplish MOT.}
	\label{fig:pipeline}
	\vspace{-0.4cm}
\end{figure*}

\section{Methodology}

We focus on processing 3D point cloud generated by LiDAR in the autonomous driving scenes. Figure \ref{fig:pipeline} is the pipeline of our proposed framework. It generally takes a pair of consecutive labelled point clouds as inputs, including old tracklets $B^{t-1}$ and new detection results $D^{t}$ in the previous and current frame severally. Finally, the framework returns the new tracklets $B^{t}$ in the updated frame. Details are illustrated in the following parts.

\subsection{Scene Preprocessing}
\label{ssec:preprocessing}

To promote efficiency and reduce the superfluous calculation, we propose to preprocess point cloud scenes frame by frame through expanding the Field of View and fitting the ground plane.

\begin{figure}[ht]
	\begin{center}
		\includegraphics[width=1.0\linewidth]{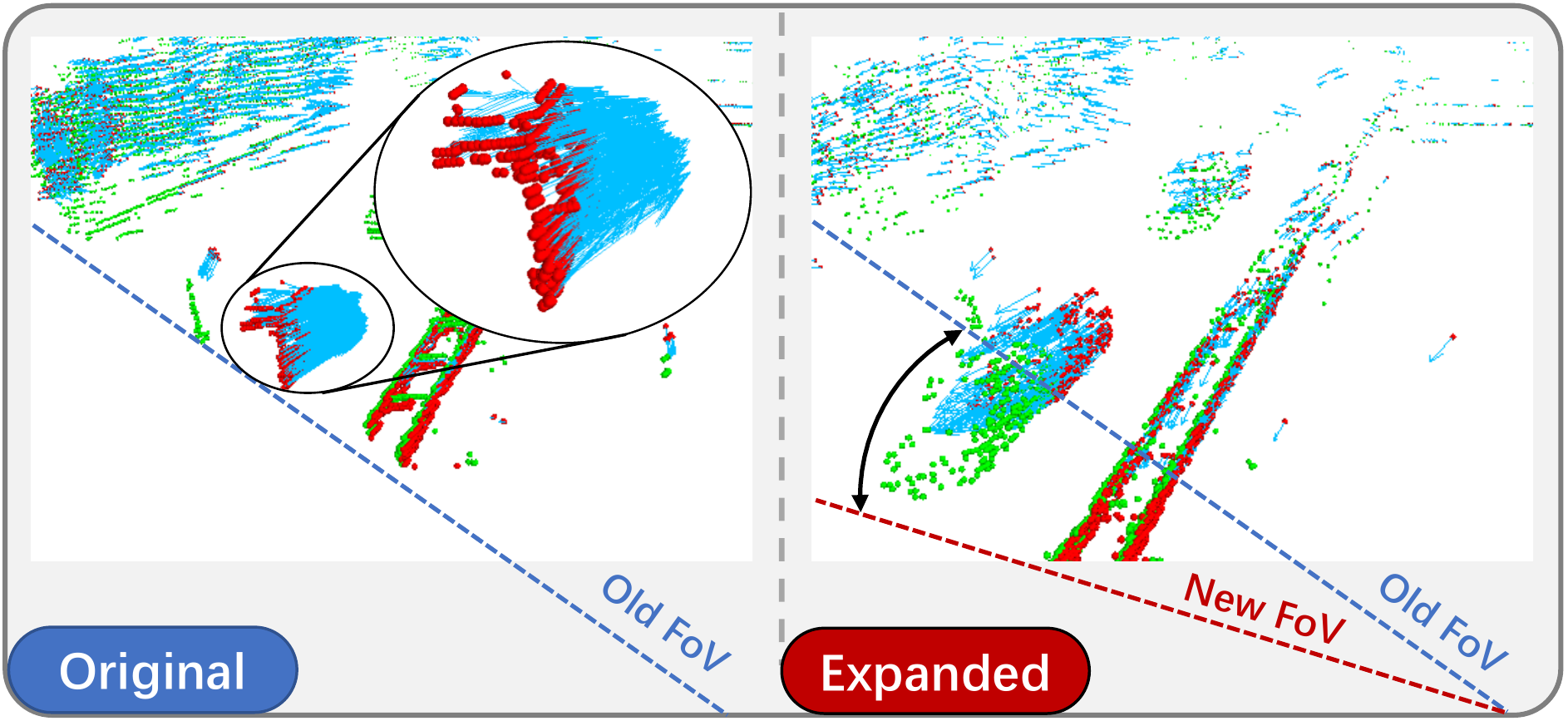}
	\end{center}
	\vspace{-2mm}
	\caption{In the original version, we notice that a car has left in the current frame, so there is no point representing it. It is a normal phenomenon that an object leaves the scene, but its vanishment would mislead corresponding points in the previous frame to a wrong scene flow estimation. To moderate this problem, we expand the FoV to let points in the current frame guide the Motion Estimation process.}
	\vspace{-0.4cm}
	\label{fig:expand}
\end{figure}

\smallskip
\noindent
\textbf{Expanded Field of View. } As the unified evaluation procedure only cares about results within the camera's Field of View (FoV), we first use the calibration relationship between the sensors to transfer the point cloud coordinates into the camera coordinate system and then reserve point clouds within the FoV. As predicting the birth and death of the objects is a crucial part of the tracking task (details in Section~\ref{ssec:tracking}), we expand the FoV to a certain range, which allows our framework to procure more information in the edge of original FoV between the previous and current frame. It helps improve the inference performance of the scene flow estimation network. Thus, it benefits the birth/death check further. Our experiments find that this proceeding is necessary. See Figure~\ref{fig:expand}.

\smallskip
\noindent
\textbf{Ground plane fitting. } As we know, points on the ground plane constitute the majority of the point cloud, and it is not necessary to concern the motion of the ground in our mission. Therefore, their removal will promote the sampling probability of potential moving objects and make the scene flow estimation more accurate (refer to Section \ref{ssec:estimation}). Various algorithms can be used in this proceeding, such as geometric clustering~\cite{zermas2017fast}, fitting plane normal vector by Random Sample Consensus (RANSAC)~\cite{behl2019pointflownet} and learning-based inference~\cite{hu2020randla}. Here we use the method of~\cite{hu2020randla}. Note that we only label the ground plane instead of directly removing it, as the front-end object detector may use raw point clouds with the ground information.

\subsection{Motion Estimation}
\label{ssec:estimation}

In this module, we leverage a learning-based framework with scene flow estimation for predicting tracklets. The key insight behind this is that 3D motion information of the entire scene is embedded in the scene flow stream regardless of object categories, and the existing learning-based scene flow estimation method has strong generalization ability and high robustness facing various scenarios. This operation modifies the predictions of old tracklets from the previous frame and further guides the following data association as well.

\smallskip
\noindent
\textbf{Data presentation. } Our work adopts FlowNet3D~\cite{liu2019flownet3d} as the scene flow estimation network that can be replaced flexibly by other potential methods. The inputs are the point clouds $P=\left\{p_{i} \mid i=1, \ldots, N_{1}\right\}$ and $Q=\left\{q_{j} \mid j=1, \ldots, N_{2}\right\}$ from the previous and current frame, where $N_1$ and $N_2$ are the quantities of points in each frame. $p_{i}$ and $q_{j}$ represent geometric and additional information of every single point as follows:
\begin{equation}
        \begin{aligned}
        p_{i} &=\left\{v_{i}, f_{i}\right\} (i = 1, \ldots, N_{1}), \\ 
        q_{j} &=\left\{v_{j}, f_{j}\right\} (j = 1, \ldots, N_{2})  
        \end{aligned}
\end{equation}
        
        \noindent
$v_{i}, v_j \in \mathbb{R}^{3}$ stand for point cloud coordinates $X,Y,Z$, while $f_{i}, f_{j} \in \mathbb{R}^{c}$ stand for color, intensity and other features (optional). The output is a bunch of scene flow vectors $F\in\mathbb{R}^{N_1 \times 3}$. The order and quantity of these vectors are consistent with the ones of points in the previous frame.

\smallskip
\noindent
\textbf{Estimation procedure. }
The aim of the scene flow estimation network is to predict locations where points in the previous frame will appear in the next one. First, the network uses \textit{set conv} layers to encode the high-dimensional features of the points randomly selected from point clouds in the two frames respectively. After this, it mixes the geometric feature similarities and spatial relationships of the points through the flow embedding layer and another \textit{set conv} layer to generate embeddings containing 3D motion information. Finally, through its \textit{set upconv} layer up-sampling and decoding the embeddings, we can obtain the scene flow of the corresponding points containing abundant motion information between two scenes~\cite{liu2019flownet3d}.

\subsection{Motion Tracking}
\label{ssec:tracking}
This module contains the back-end tracking procedures. It explicitly expresses the movement tendencies of objects from the extraction of the point-wise motion consistency.

\smallskip
\noindent
\textbf{Instance prediction. }  The goal is to estimate the movement of every tracklet in the previous frame. According to the result given by the detector and old tracklets, we can allocate a unique instance ID to each object, separating them from each other. After obtaining the scene flow of the whole scene between frames, we can predict the approximate locomotive translation and orientation of each object according to instance IDs by calculating the linear increment prediction of each tracklet $(\Delta x_{n}, \Delta y_{n}, \Delta z_{n})$ and the angular increment prediction $\Delta \theta_{n}$ which can be individually obtained from Equation~\ref{equal:mean} and the constant angular velocity model. 

\begin{equation}\label{equal:mean}
(\Delta x_{n}, \Delta y_{n}, \Delta z_{n})=\frac{1}{N} \sum_{c=1}^{N}{F_{c}^{n}}
\end{equation}
${F_{c}^{n}}$ means the scene flow vector of the $c$th point belonging to the corresponding tracklet and $N$ is the quantity of points. The final combined increment is called \textit{offset} represented as:
\begin{equation}
\begin{aligned}
O &=\left\{o_{n} \mid n=1, \ldots, T\right\}, \\ 
o_{n} &=\left\{(\Delta x_{n}, \Delta y_{n}, \Delta z_{n}), \Delta\theta_{n}\right\} 
\end{aligned}
\end{equation}
where $T$ is the quantity of tracklets in the previous frame.

\smallskip
\noindent
\textbf{Data association. } In every frame, the state of whole tracklets and the bounding box of each tracklet can be presented as
\begin{equation}
\begin{aligned}
& B^{t-1} = \left\{b_{n}^{t-1} \mid n = 1, \cdots, T\right\},\\ 
& b_{n}^{t-1} = \left\{x_{n}^{t-1},y_{n}^{t-1},z_{n}^{t-1},l_{n}^{t-1},w_{n}^{t-1},h_{n}^{t-1},\theta_{n}^{t-1}\right\}
\end{aligned}
\end{equation}including location of the object center in the 3D space $(x, y, z)$, object 3D size $(l, w, h)$ and heading angle $\theta$. The state will be predicted as $B_{pre}^{t-1}$ by \textit{offset} 
\begin{equation}
        (X_{pre}^{t-1}, Y_{pre}^{t-1}, Z_{pre}^{t-1})=(X, Y, Z)+(\Delta X, \Delta Y, \Delta Z)
\end{equation}
\begin{equation}
\Theta_{pre}^{t-1}=\Theta+\Delta\Theta
\end{equation}

\smallskip
\noindent
Then we calculate a similarity score matrix with $B_{pre}^{t-1}$ and the new object detection $D^{t}=\left\{d_{n}^{t} \mid n = 1, \cdots, M\right\}$ in the $t$ frame. The matching matrix $S\in{\mathbb{R}^{T \times M}}$ produces scores for all possible matches between the detections and tracklets, which can be solved in polynomial time by Hungarian matching algorithm~\cite{kuhn1955hungarian}.

\smallskip
\noindent
\textbf{ID propagation. } Most methods lying in the joint detection and tracking paradigm share a common procedure: they first obtain the link information between detection results in the previous and current frame and then directly propagate tracking IDs according to this prior information without bells and whistles. We use this strategy without introducing complicated calculations in the new tracklets generation process, but propagate IDs according to the instance matching relationship generated by data association. That is, we fully trust and adopt the detection results in the next frame. If we fail to align an old tracklet to its corresponding detection, we will check its age and decide whether to keep it alive.

\smallskip
\noindent
\textbf{Birth/Death check. } This inspection serves to determine whether a tracklet meets its birth or death. There are two circumstances that can cause a trajectory of birth and death respectively. The normal one is that the tracked objects leave the scene or new objects enter the scene, whilst the abnormal but common one is caused by the wrong detection results from the front-end detector we use -- \textit{False Negative} and \textit{False Positive} phenomena. We handle these problems by setting a \textit{max$\_$mis} and a \textit{min$\_$det}, which individually represents the maintenance of a lost tracklet and the number of observation of a tracklet candidate.

\textit{1) Birth check:} On the one side, we treat all unmatched detection results $D^{det}$ as potential new objects entering the scene, including False Positive ones. Then we deal with this in an uniform way -- the unmatched detection result $d_{n}^{det}$ will not create a new tracklet $b_{n}^{bir}$ for $d_{n}^{det}$ until the continuous matching in the next \textit{min$\_$det} frame. Ideally, this strategy will avoid generating wrong tracklets.

\textit{2) Death check:} On the other side, we treat all unmatched tracklets $B^{mis}$ as potential objects leaving the scene, including False Negative ones. Also we deal with this in another uniform way -- when a tracklet $b_{m}^{miss}$ cannot be associated with a corresponding detection in the next frame, its age will increase and we will keep updating attributes of the bounding box unless the age exceeds \textit{max$\_$mis}. Typically, this tactic will keep a true tracklet still alive in the scene but cannot be found a match due to a lack of positive detection results.

\begin{figure*}[t]
        \begin{center}
                \includegraphics[width=1.0\linewidth]{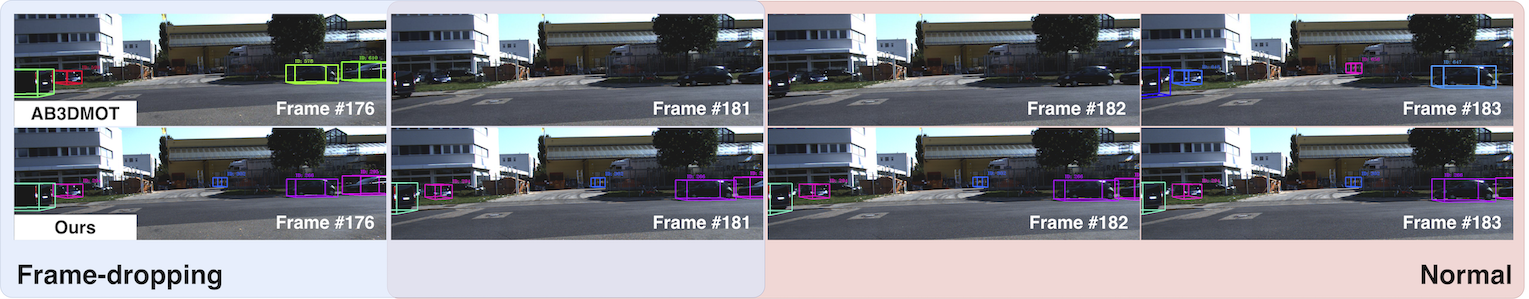}
        \end{center}
        \caption{LiDAR Frame-dropping circumstance in KITTI sequence 01. For providing a better viewing experience, we show Frame 176 to Frame 183 from the camera, which are consistent with the ones from the LiDAR. Notice that AB3DMOT dies at Frame 181 and respawns at Frame 183.}
        \label{fig:diff}
        \vspace{-0.35cm}
\end{figure*}

\begin{table*}[!h]
\caption{Performance of Car on the KITTI 10Hz val dataset using the \textbf{3D} MOT evaluation metrics.}
\centering
\resizebox{\textwidth}{!}{
\begin{tabular}{@{}lllccccccccc@{}}
\toprule
Method & Algorithm & Input Data & Matching criteria & \textbf{sAMOTA}$\uparrow$ & AMOTA$\uparrow$ & AMOTP$\uparrow$ & MOTA$\uparrow$ & MOTP$\uparrow$ & IDS$\downarrow$ & FRAG$\downarrow$\\
\midrule
mmMOT~\cite{zhang2019robust} & Learning & 2D + 3D 
        & $\text{IoU}_{\text{thres}}$ = 0.25 & 70.61 & 33.08 & 72.45 & 74.07 & 78.16 & 10 & 55\\
& & & $\text{IoU}_{\text{thres}}$ = 0.7 & 63.91 & 24.91 & \textbf{67.32} & 51.91 & 80.71 & 24 & 141 \\ 
                                        
FANTrack~\cite{baser2019fantrack} & Learning & 2D + 3D 
        & $\text{IoU}_{\text{thres}}$ = 0.25 & 82.97 & 40.03 & 75.01 & 74.30 & 75.24 & 35 & 202 \\
& & & $\text{IoU}_{\text{thres}}$ = 0.7 & 62.72 & 24.71 & 66.06 & 49.19 & 79.01 & 38 & 406 \\

AB3DMOT~\cite{weng20203d} & Filter & 3D 
        & $\text{IoU}_{\text{thres}}$ = 0.25 & \textbf{93.28} & \textbf{45.43} & \textbf{77.41} & \textbf{86.24} & 78.43 & \textbf{0} & \textbf{15} \\
& & & $\text{IoU}_{\text{thres}}$ = 0.7 & 69.81 & 27.26 & 67.00 & 57.06 & \textbf{82.43} & \textbf{0} & \textbf{157} &  \\
\midrule
FlowMOT (Ours) & Hybrid & 3D 
        & $\text{IoU}_{\text{thres}}$ = 0.25 & 90.56 & 43.51 & 76.08 & 85.13 & \textbf{79.37} & 1 & 26 \\
& & & $\text{IoU}_{\text{thres}}$ = 0.7 & \textbf{73.29} & \textbf{29.51} & 67.06 & \textbf{62.67} & 82.25 & 1 & 249 &  \\
\bottomrule
\end{tabular}}
\vspace{-0.4cm}
\label{tab:3dcomparison}
\end{table*}

\begin{table}[t]
        \caption{Performance of Car on the KITTI 5Hz (high-speed) val dataset.}
        \vspace{-0.2cm}
        \centering
        \resizebox{\hsize}{!}{
        \begin{tabular}{@{}lccccc@{}}
        \toprule
        Method & Matching criteria & \textbf{sAMOTA}$\uparrow$ & AMOTA$\uparrow$ & AMOTP$\uparrow$ & MOTA$\uparrow$ \\
        \midrule
        AB3DMOT~\cite{weng20203d} & $\text{IoU}_{\text{thres}}$ = 0.25 & 82.98 & 36.83 & 69.77 & 74.55 \\
                        & $\text{IoU}_{\text{thres}}$ = 0.7 & 56.71 & 18.96 & 58.00 & 45.25 \\
        FlowMOT (Ours)    & $\text{IoU}_{\text{thres}}$ = 0.25 & \textbf{87.30} & \textbf{40.34} & \textbf{74.22} & \textbf{80.11} \\
                        & $\text{IoU}_{\text{thres}}$ = 0.7 & \textbf{70.35} & \textbf{27.23} & \textbf{65.20} & \textbf{59.72} \\
        \bottomrule
        \end{tabular}}
        \vspace{-0.4cm}
        \label{tab:drop1}
        \end{table}

\section{Experiments}

\subsection{Experimental protocols:}\label{imple}

\smallskip
\noindent
{\bf Evaluation Metrics.}
We adopt the 3D MOT metrics firstly proposed by AB3DMOT~\cite{weng20203d}, which are more tenable on evaluating the performance of 3D MOT systems than the ones in CLEAR~\cite{bernardin2008evaluating} focusing on 2D performance. The primary metric used to rank tracking methods is also called MOTA, which incorporates three error types: False Positives (FP), False Negatives (FN) and ID Switches (IDS). The difference lies in the calculation ways of these factors modified from 2D IoU to 3D IoU, and we match the 3D tracking results with ground truth directly in 3D space. Meanwhile, MOTP and FRAG are also used as parts of the metrics.

In addition, AB3DMOT illustrates that the confidence of detection results can dramatically affect the performance of a specific MOT system (details are delivered in~\cite{weng20203d}). It proposes AMOTA and AMOTP (average MOTA and MOTP) to handle the problem that current MOT evaluation metrics ignore the confidence and only evaluate at a specific threshold.

\begin{equation}
\label{equal:AMOTA}
\text{AMOTA} = \frac{1}{L} \sum_{r \in \{\frac{1}{L}, \cdots, 1\}} \text{MOTA}_r,
\vspace{-0.1cm}
\end{equation}
where $r$ is a specific recall value (confidence threshold) and $L$ is the amount of recall values. AMOTP is in the similar format. To make the value of the AMOTA range from 0\% up to 100\%, it scales the range of the $\text{MOTA}_r$ by:
\begin{equation}
\text{sMOTA}_{r} =\max \left(0,\frac{1}{r} \text{MOTA}_r\right)
\end{equation}
The average version sAMOTA can be calculated via Equation~\ref{equal:AMOTA} which is the most significant one among all metrics.

\smallskip
\noindent
{\bf Dataset settings. }
We evaluate our framework on the KITTI MOT dataset~\cite{Geiger2012CVPR}, which provides high-quality LiDAR point clouds and MOT ground truth. Since KITTI does not provide the official train/val split, we follow~\cite{scheidegger2018mono} and use sequences 01, 06, 08, 10, 12, 13, 14, 15, 16, 18, 19 as the val dataset and other sequences as the training dataset, though our 3D MOT system does not require training on this dataset. We adopt the evaluation tool provided by AB3DMOT as KITTI only supports 2D MOT evaluation. For the matching principle, we follow the convention in KITTI 3D object detection benchmark and use 3D IoU to determine a successful match. Specifically, we use 3D IoU threshold $\text{IoU}_{\text{thres}}$ of 0.25 and 0.7 representing a tolerant criterion and a strict one respectively for the Car category and 0.25 for the Pedestrian category. To evaluate the robustness, we generate a subset by downsampling the val dataset from 10Hz to 5Hz to simulate high-speed scenes as parts of variable-speed scenarios.

\begin{figure*}[t]
	\begin{center}
		\includegraphics[width=1.0\linewidth]{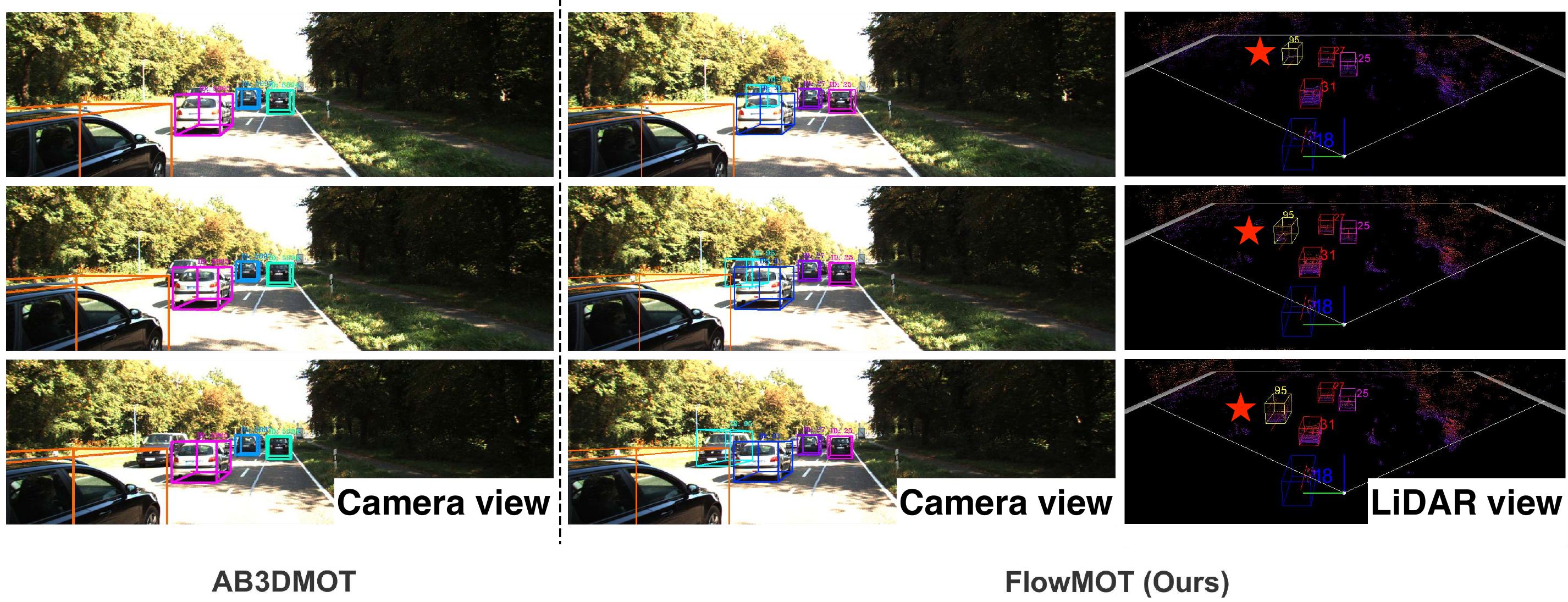}
	\end{center}
	\caption{Qualitative results of FlowMOT on our 5Hz split. The first column shows the predicted results produced by AB3DMOT~\cite{weng20203d}, whilst the second and third column is our results both in the camera view and LiDAR view. Note that AB3DMOT loses the target car represented by a red star in the third column, but we can track it successfully. (Best viewed with zoom-in.)}
	\vspace{-0.3cm}
	\label{fig:exhibition}
\end{figure*}
\smallskip
\noindent
{\bf Implementation details.}
\label{ssec:sampling}
For choosing the front-end detector, we adopt a 3D detection method purely based on the point cloud. In order to compare with other methods fairly, we use PointRCNN~\cite{shi2019pointrcnn} as the detection module, and off-the-shelf detection results are consistent with the ones conducted in \cite{weng20203d}.

Recently, several 3D scene flow approaches~\cite{liu2019flownet3d,gu2019hplflownet,liu2019meteornet} have been proposed and showed terrific performance, which can be easily integrated with our \textit{Motion Estimation} module. In our experiments, we adopt FlowNet3D~\cite{liu2019flownet3d} and deploy it on an Nvidia Titan Xp GPU. Since the point clouds in KITTI MOT dataset lack the ground truth labels of the scene flow, we keep the training strategy of the network consistent with the one conveyed in its original paper -- using FlyingThings3D~\cite{mayer2016large} as the training dataset. Although the data set is a synthetic data set, FlowNet3D has a strong generalization ability and can adequately adapt to the LiDAR-based point cloud data distribution in KITTI. For specific, we randomly sample 6000 points for each frame resulting in 12000 points (two adjacent frames) as the network input. Meanwhile, we notice that~\cite{mittal2020just} proposes to train FlowNet3D in an unsupervised manner, making the training process of the network no longer bound by the lack of ground truth, which shows more potential and may further improve the performance of our framework. 


With respect to the designing of \textit{Motion Tracking} module, we use the same parameters in AB3DMOT, aiming to compare our performance without the effect caused by other factors. We use $\text{IoU}_{min}$ = 0.01 in the match principle of the data association part. In the birth and death memory module, \textit{max$\_$mis} and \textit{min$\_$det} are set to 2 and 3 respectively.

\subsection{Experimental Results:}\label{compare}

\begin{table}[t]
\caption{Performance of Pedestrian with $\text{IoU}_{\text{thres}}$ = 0.25 on the KITTI 10Hz and 5Hz val dataset.}
\vspace{-0.1cm}
\centering
\resizebox{\hsize}{!}{
\begin{tabular}{@{}lccccc@{}}
\toprule
Method & Frame rate & \textbf{sAMOTA}$\uparrow$ & AMOTA$\uparrow$ & AMOTP$\uparrow$ & MOTA$\uparrow$ \\
\midrule
AB3DMOT~\cite{weng20203d} & 10Hz & 75.85 & 31.04 & \textbf{55.53} & \textbf{70.90} \\
            & 5Hz & 62.53 & 20.91 & 44.50 & 57.90 \\
FlowMOT (Ours)    & 10Hz & \textbf{77.11} & \textbf{31.50} & 55.04 & 69.02 \\
            & 5Hz & \textbf{74.58} & \textbf{29.73} & \textbf{53.23} & \textbf{66.97} \\
\bottomrule
\end{tabular}}
\vspace{-0.5cm}
\label{tab:drop2}
\end{table}

\smallskip
\noindent
\textbf{Quantitative results. }
The quantitative experiment results are divided into two parts -- the tracking result on the KITTI val dataset at the normal frame rate and the result with reduced frame rate to imitate the high-speed scenes. The most critical indicator is sAMOTA, as MOTA has some inherent drawbacks.

\textit{1) Normal scenes: }We compare against state-of-the-art methods that are both learning-based and filter-based. We conclude the results in Table~\ref{tab:3dcomparison} and Table~\ref{tab:drop2}. Our system can outperform mmMOT (ICCV$'$19) and FANTrack (IV$'$19) to a large margin and meanwhile achieve a competitive result with AB3DMOT (IROS$'$20). For the Car category, we consider that the reason why we are slightly inferior to AB3DMOT on tolerant $\text{IoU}_{\text{thres}}$ is that KITTI lacks scene flow ground truth and FlowNet3D is unable to be fine-tuned adequately. When $\text{IoU}_{\text{thres}}$ becomes stricter, our ID propagation strategy starts to be effective as raw detection results are more accurate toward ground truth bounding boxes than filter-updated tracklets. Thus, our framework can outperform AB3DMOT. For the Pedestrian category, Table~\ref{tab:drop2} shows ours can be superior to AB3DMOT. The reason is that the motion style of Pedestrian is different from Car, and thus the hyperparameter settings for Pedestrian in AB3DMOT should not consist with the ones for Car. However, due to the inherent problem of constant covariance matrix, AB3DMOT should repeatedly adjust matrix to maintain its performance across various categories. In contrast, our method can conduct one-off tracking procedure for all categories, as we achieve a various motion model by estimating scene flow.

\textit{2) High-speed scenes: } In practice, the various-speed motion of vehicles is common and the frame rate of sensors is usually different from each other. Thus, it is important to evaluate robustness. We first observe that AB3DMOT shows a performance descent in sequence 01 as Frame 177 to Frame 180 are missing, making sequence 01 a different frame rate compared with other sequences, as shown in Figure~\ref{fig:diff}. It implies that the single constant motion used in AB3DMOT may obtain more failures on the various-speed scenarios and cannot handle the frame-dropping problem which is common in the practical daily usage of sensors. Inspired by this, we further imitate high-speed scenes, which belong to various-speed scenarios, by only keeping the even-numbered frames in the dataset equivalent to reducing the frame rate to 5Hz. The quantitative comparison with AB3DMOT proves our statement and it is summarized in Table~\ref{tab:drop1} and Table~\ref{tab:drop2}. 


\smallskip
\noindent
\textbf{Qualitative results. }
Figure~\ref{fig:exhibition} shows the performance of our method and AB3DMOT facing the various frame rate scenarios. It shows that AB3DMOT cannot handle a lower frame rate representing high-speed or frame-dropping circumstances because the parameters in the covariance matrix used in this filter-based method are only suitable for a specific frame rate, and it fails when the frame rate changes. Our method takes advantage of the robustness and generalization of the learning-based algorithm and can better adapt to this scenario than AB3DMOT.

\section{Discussion}
Scene flow has shown its advantages in predicting objects' motion and further improve the robustness of MOT, but there are a few limitations we are still facing. Firstly, due to the data distribution of LiDAR-based point cloud, the density of the point cloud changes with the distance. This phenomenon make FlowNet3D's estimation relatively inaccurate at a distance, as FlowNet3D is mainly trained on dense depth images. Thus, tracklets in the distance are not well associated causing tracking failure. One solution is the variable density sampling, which can increase more attention on farther points and keep less closer points. Secondly, at present, networks used for scene flow estimation are always designed to be large, and therefore they tend to occupy much memory usage of GPU. How to produce more lightweight network is a future direction researchers can study. 

\section{Conclusion}
This paper has presented a tracking-by-detection 3D MOT framework, which is the first to propose to predict and associate tracklets by learning-based scene flow estimation. We take advantage of the traditional matching method and the deep learning approach to avoid the problems caused by cumbersome manual hyperparameter adjustment and constant motion model existing in filter-based methods. The experiments show that our method is more accurate than state-of-the-art end-to-end methods and obtain competitive results against filter-based methods. Moreover, our framework can succeed to process challenge scenarios where the accurate filter-based method may break down. Despite some limitations, we hope our work can be a prototype to inspire researchers to explore related fields.


\addtolength{\textheight}{-7.2cm}   

\bibliographystyle{IEEEtran}
\bibliography{root}

\end{document}